\documentclass{article}

\usepackage[preprint]{neurips_2023}

\usepackage[ruled,vlined]{algorithm2e}
\usepackage{amsmath}
\usepackage{amssymb}
\usepackage{bm}             
\usepackage{tabularx}
\usepackage{ragged2e}
\usepackage{graphicx}
\usepackage{caption}
\usepackage{subcaption}
\usepackage{booktabs}
\usepackage{url}
\usepackage{hyperref}       
\usepackage{mdframed}       
\usepackage{cleveref}       

\begin{document}


\title{Topological quantification of ambiguity in semantic search}

\author{%
    Thomas R. Barillot\\
    Avantia \\
    London, United Kingdom \\
    \And
    Alex De Castro \\
    Imperial College London\\
    London, United Kingdom \\
}




\maketitle

\begin{abstract}
We studied how the local topological structure of sentence-embedding
neighborhoods encodes semantic ambiguity.  Extending ideas that link
word-level polysemy to non-trivial persistent homology, we generalized the concept
to full sentences and quantified ambiguity of a query in a semantic search process with two persistent homology metrics: the 1-Wasserstein norm of \(H_{0}\) and the maximum loop
lifetime of \(H_{1}\).

We formalized the notion of ambiguity as the relative presence of semantic domains or topics in sentences. We then used this formalism to compute "ab-initio" simulations that encode datapoints as linear combination of randomly generated single topics vectors in an arbitrary embedding space and demonstrate that ambiguous sentences separate from unambiguous ones in both metrics.

Finally we validated those findings with real-world case by investigating on a fully open corpus comprising Nobel Prize
Physics lectures from 1901 to 2024, segmented into contiguous,
non-overlapping chunks at two granularity: \(\sim\!250\) tokens and \(\sim\!750\) tokens.  We tested embedding with four publicly available models. Results across all models reproduce simulations and remain stable despite changes in embedding architecture. 

We conclude that persistent homology provides a model-agnostic signal of semantic discontinuities, suggesting practical use for ambiguity detection and semantic search recall.
\end{abstract}

\textbf{Keywords:} Semantic Search, Topological Data Analysis, Ambiguity Quantification




\section{Introduction}

In the field of natural language processing, semantic matching between sentence embeddings  often employs metrics such as Euclidean distance, dot product or cosine similarity. However, the effectiveness of these measures can be undermined by the fact that the manifold constructed by sentence embeddings is unlikely to be globally smooth, and it may not be locally smooth either. The nature of the semantic manifold, often hypothesized to be low-dimensional based on empirical evidence and intuition, remains elusive yet critical. Literature contains attempts to smooth manifold representation for words and sentences (\cite{hasan-curry-2017-word,Kayal2021,chu2023rsbert}) but they can be seen as fine-tuning exercises that are by construction limited to a certain domain.
The elusive structure of the semantic manifold and its probable discontinuous nature are particularly problematic for queries retrieval tasks, which may consist in multi-factual or too vague (ambiguous) questions. Recent studies have leveraged Topological Data Analysis (TDA, \cite{carlsson2021}) to interpret the polysemic nature of words (\cite{jakubowski2020topology}), indicating local discontinuities within an otherwise smooth word embedding manifold. Building on the concept of polysemy in words, we promote a working and computational definition of ambiguity in sentences based on our experience with vector search and RAG (\cite{gao2024retrievalaugmented}) systems in tackling disambiguation problems.

Polysemy refers to a word having multiple meanings, which occurs when a word belongs to more than one semantic domains which are mutually exclusive. In other words, given a unique polysemic word, inferring any associated semantic domain (or \textbf{topics}) is equiprobable.
Similarly, we examine the \textit{relative ambiguity} in sentences within a corpus.

\textbf{Intuitive definition:} An polysemic and therefore ambiguous sentence is formally defined as a sentence whose embedding satisfies the conditions of connecting to less topics within its corpus than the minimal number of connections established by other sentences, or not fitting neatly into any predefined topic of the corpus.

\textbf{Toy Example:}
\begin{itemize}
    \item Consider sentences involving the word \textit{crane}:
          \begin{itemize}
              \item Sentence 1: "He craned his neck to see the construction site."
              \item Sentence 2: "The crane lifted the heavy beams effortlessly."
              \item Sentence 3: "A crane flew over the lake at sunset."
          \end{itemize}
          Topics might include \{\textit{action}, \textit{machine}, \textit{bird}\}. Sentence 1 connects to \{\textit{action}\}, Sentence 2 to \{\textit{machine}\}, and Sentence 3 to \{\textit{bird}\}.
    \item To illustrate polysemy:
          \begin{itemize}
              \item "I saw a crane."
          \end{itemize}
          This sentence is ambiguous as it lacks context to ground the word "crane" in a particular context.
    \item A multi-factual query would be:
          \begin{itemize}
              \item "He craned his neck to watch the crane lift near the crane."
          \end{itemize}
          This sentence employs all three topics, making it a multi-factual example where the word "crane" connects to \{\textit{action}, \textit{machine}, \textit{bird}\} simultaneously.
\end{itemize}

A corpus \( \mathcal{C} \) comprises \( n \) sentences, or text chunks, represented as \( \mathcal{C} = \{c_1, ..., c_n\}\), with each sentence potentially fitting into one or more of \( m \) identified topics \( S_{\mathcal{C}} = \{s_1, ..., s_m\} \), where \( m \leq n \). Also for every chunk \(c_j\) we will associate a subset of topics \(S_{c_j}\).

We now consider a (typically new) sentence or chunk \( c' \) with a topic set \(S_{c'}\) polysemic relative to a corpus \(\mathcal{C}\) if it meets the following \textbf{working definition}:

\begin{equation*}
    |S_{c'}| < \min_{c}(|S_{c}|) \quad \text{with} \quad S_{c'} \subset S_{\mathcal{C}}
\end{equation*}
It contains less topics than the sentence in the corpus with the fewest number of topics. On the contrary, the sentence is a composition of topics (multi-factual) belonging to \(S_{\mathcal{C}}\) if:
\begin{equation*}
    S_{c} \subset S_{c'} \quad \forall  S_{c} \subset  S_{\mathcal{C}}
\end{equation*}

Finally it is unrelated to \(\mathcal{C}\) if it does not belong to any recognized semantic domain:
\begin{equation*}
  S_{c'} \cap S_{\mathcal{C}} = \{\emptyset\} 
\end{equation*}

Notice that in the latter case \(c'\) does not pertain to the original corpus \(\mathcal{C}\). In applications, new chunks \(c'\) represent unseen incoming queries and in the vast majority of cases they meet polysemic or multi-factual conditions.

\textbf{We do not claim that this is the only possible definition}. We welcome feedback from the computational NLP community to understand if there are other practical and computable definitions, and whether they can be proven mathematically or experimentally to be equivalent.

The generalization from word polysemy to sentence ambiguity raises research questions that we propose to explore:\\
\begin{mdframed}
    \begin{enumerate}
        \item Does an ambiguous sentence embedding display a distinctive geometric and topological signature within its neighborhood, as observed with polysemic word embeddings?
        \item And if so, does the signature depend on a specific training dataset or embedding model architecture?
    \end{enumerate}
\end{mdframed}

It is important to note that clustering of sentence embeddings has been successfully explored with Local Density Approximation or more recently BERTopic (\cite{grootendorst2022bertopic}) to represent  topics. Therefore it follows that relative ambiguity reflects the fact that incoming queries may be associated with a subset of pre-existing topics present in the corpus. This can have a degrading impact on the precision of RAG systems that depend on clear semantic context for providing answers with high utility for end users. On the other hand, a multi-factual query might be related to multiple elements in the corpus and retrieval ranking system will fail to address complete retrieval of the information (low recall).
Our goal is to be able to quantify and observe the degree of ambiguity in a sentence \(c\) in relation to the corpus \(\mathcal{C}\), so we can mitigate or correct the effects of the additional relative ambiguity when retrieving information or generating new content based on an existing context window.

To address these questions we translated the formal working definitions above into an "ab-initio" simulation relying solely on the \textbf{linear representation hypothesis} where topics are represented by a set of orthogonal feature vectors encoded in the embedding space and where abstract sentence construction is a linear combination of those features (details in appendix \ref{appendix:recursive-homology}). This hypothesis as unique axiom for our simulation is strongly motivated by the works of Anthropic on information encoding description (\cite{Elhage2022}) and recent researches that empirically evidence linearly separable representations of semantic domains in large language models
hidden states (\cite{Saglam2026}). We designed topological features using Topological Data Analysis toolbox and evaluated their sensitivity to the presence or absence of multiple topics.

We then conducted a retrieval experiment on an open-access corpus consisting of Nobel Prize Physics lectures delivered from 1901 to 2024. The lectures were tokenized and chunked in two granularities (250-token and 750-token windows). We then tested queries that either relate to a single or multiple topics covered in the corpus of answers.

Key findings include the exploration of the relative nature of sentence ambiguity, demonstrating how the number of concepts contained in the query and corpus affect the occurrence of local manifold discontinuities. We believe our findings reinforce the mechanistic approach of information encoding in transformer based LLMs and that it can be used to design strategies to improve retrieval in RAG systems optimizing the precision-recall trade-off.


\section{Related works}

Ambiguity created by similarity-metric limitations propagate into retrieval errors in RAG pipelines. Detecting and mitigating those limitations are objects of extensive researches. Several recent works approach this problem:
CRAG (\cite{yan2024corrective}) explicitly categorize in \{"Correct","Ambiguous","Incorrect"\} a retrieved chunk of text for a given query using a T5-large fine tuned model and subsequently triggers different actionables. That work reports a whole retrieval augmentation pipeline and is an elegant solution but does not give any insights about explainability of incorrect retrievals in terms of ambiguity and is domain dependent as it relies on a fine-tuned language model. SELF-RAG (\cite{asai2023selfrag}) also evaluates the relevance of each retrieved chunks given a specific query but by directly using the Large Language Model at generation step. This model is fine-tuned to generate additional relevance tokens to the output in order to trigger generation or a new retrieval loop. As for CRAG, this works focuses on producing flags to iterate the retrieval process but does not interrogate the notion of ambiguity or relevance.\\
It is also worth mentioning that research groups also question widely used scoring methods such as cosine similarity as unique metric for semantic similarity (\cite{zhou2022problems,Steck2024cosine},) mostly due to words frequency induced biases in training (\cite{wannasuphoprasit2023solving}). These questions call for an exploration of additional metrics to characterize the semantic search results. In investigating queries ambiguity, we distinctively focus on an understanding of domain-specific polysemy using topological and geometric methods. Very recent work defined geometric based metric: "the semantic grounding index" by computing angular distance between embeddings with the objective to detect whether large language model answers were similar to the context provided (likely valid answer) or just to the query (likely hallucination) in a RAG framework (\cite{Marin2025}). 

We also link our work to the LLM explainability research domain . In particular we are interested in framing the problem in terms of linear representation hypothesis that consists in modeling information features as quasi-orthogonal directions in the model hidden states dimensions. Anthropic published empirical results converging to such a description of information encoding (\cite{Elhage2022}). They also showed, using sparse auto encoders (SAE) (\cite{Bricken2023},\cite{Templeton2024})in their models that such features can be represented by low dimensional manifolds and used to steer generative LLMs outputs in a controlled fashion.Those works have opened the door to research on the features manifold geometries (\cite{JEngel2025}],\cite{AModell2025}).

On a practical aspect, current algorithms focus on ranking retrieved documents to optimize precision and approximate recall through metrics such as Mean Reciprocal Ranking  for a predefined top-k documents (MMR@\(k\)) (\cite{Robertson2009,Formal2021,Santhanam2022}). Little to no attention is given to optimize the true recall of all meaningful datapoints (ie: inferring \(k\)) and the problem is delegated either:
\begin{itemize}
    \item To the documents chunking strategy which can be difficult to validate and is dataset or domain dependent. Moreover chunking cannot account for long range correlation in a document.
    \item To an arbitrary cut-off in the ranked results which cumulates dataset dependency and query dependency.
\end{itemize}
We intend to demonstrate that our working definition of ambiguity is a measurable feature of language and that, in the context of LLM representation of language, it that can compensate the shortcomings of similarity methods.

\section{Methodology}
\subsection{Persistent homology as a metric.}
Persistent homology provides a way to transform the geometric structure of a point cloud into quantitative features. 
Given a set of points, we gradually increase a connectivity radius and track how the topology of the neighborhood evolves.

At small radii, all points are isolated. 
As the radius increases, nearby points connect and form clusters. 
Further increasing the radius may cause clusters to merge or create loop-like structures before eventually collapsing into a single connected component.

The zeroth homology group, $H_0$, records the birth and merging of connected components. 
Its persistence values therefore encode how connectivity emerges across scales and provide a multiscale measure of neighborhood fragmentation.

The first homology group, $H_1$, records the appearance and disappearance of loops. 
Persistent loops indicate the presence of distinct substructures separated by geometric gaps within the neighborhood.

By summarizing these persistence values (through Wasserstein norms for $H_0$ and maximum lifetimes for $H_1$), we obtain numerical descriptors that reflect the multiscale organization of the query neighborhood. 
These descriptors serve as topology-aware metrics that complement standard similarity-based measures.

\subsection{Evaluating query neighborhood}

The features employed in our experiment are derived from the relative positions of the query embedding vector \(\vec{v}_{q}\) and the embedding vectors of its \(k\) nearest neighbors in the answer corpus, denoted by \(\{\Delta \vec{v}_{iq}\}_{i=0}^{k-1}\) where \(\Delta \vec{v}_{iq} = \vec{v}_{i} - \vec{v}_{q}\).

Each neighbor of the query relative distance is represented as a scaling factor \(\varepsilon_{i}\) w.r.t the relative distance with the nearest neighbor with \(\varepsilon_{0}\) = 0.
\begin{equation}
    d(i,q) = \frac{\Delta \vec{v}_{iq}}{\Delta \vec{v}_{0q}} \equiv 1 + \varepsilon_{i}
    \label{eq:f_scale_mean}
\end{equation}

We opted for this representation as it reflects the local neighbors density which the ranking index \(k\) does not. We can then tune an arbitrary scaling factor \(\varepsilon_{1}\leq\varepsilon\leq\varepsilon_{k}\) and observe the evolution of topological features. The value of k is taken sufficiently high to allow probing both local and global neighborhood.

Topological features are constructed based on persistent homology of order 0 and 1, denoted as \(H_{0}\) and \(H_{1}\). Following the methodology presented in \cite{jakubowski2020topology}, we calculate them by first normalizing the nearest neighbor vectors:
\[
    \frac{\Delta \vec{v}_{iq}}{\|\Delta \vec{v}_{iq}\|}
\]
and applying the persistent homology algorithm, as referenced in \cite{tauzin2021giottotda}. Practically, Wasserstein distance and maximum loop lifetime are chosen to quantify \(H_{0}\) and \(H_{1}\) respectively.

\subsection{Understanding Wasserstein Distance in Topological Data Analysis}

The Wasserstein distance, also known as the earth mover's distance, quantifies the "work" required to transform one distribution into another. In topological data analysis, this measure is particularly useful when applied to persistence diagrams. The diagonal of a persistence diagram ideally represents features (often considered as noise) that appear and disappear at the same scale, thus serving as a baseline for comparison.

For a set of points in a persistence diagram that represent the death times of homology-0 features (e.g., regular lattices), the 1-Wasserstein distance can be conceptualized as the cost of moving this distribution onto the diagonal. This distribution of death times on the y-axis not only reflects the moment each feature disappears but also indicates the granularity of the lattice at which these features resolve. Each y-value, therefore, represents the radius within which a particular regular lattice ceases to exist, providing insights into the scale of data concentration and granularity.

The practical calculation of the 1-Wasserstein for \(H_{0}\) distance involves:
\begin{equation}
    W_{1}(H_{0}) = \frac{1}{N-1}\sum_{y \in H_{0}} |y - \gamma_{\perp}(y)|
    \label{eq:1-wasserstein-h0}
\end{equation}
where \(N\) is the number of points in \(H_{0}\), and \(\gamma_{\perp}(y)\) is the y-coordinate of the orthogonal projection of each point onto the diagram's diagonal. This calculation measures how far each point deviates from the baseline (diagonal), it implicitly weighs the features by their granularity. Each distance moved in the calculation represents a weighted adjustment of the data’s granularity, reflecting varying scales of data clustering and feature resolution.


\subsection{Maximum loop lifetime}

The next homology order \(H_1\), which captures 1-dimensional loops in the data, is more sensitive to noise and therefore the usage of equation \ref{eq:1-wasserstein-h0} would artificially reduce long-lived loops weights, therefore we opted for representing the longest-lived loop in the persistence diagram:

\begin{equation}
    LT_{max}(H_1) = \sup_{(x,y) \in H_1} [|y - \gamma_{\perp}(y)|]
    \label{eq:h1-lifetime}
\end{equation}

This increased sensitivity of \(H_1\) features to noise arises from the complexity and the intricate structures they capture. Noise can introduce spurious loops with short lifespans, affecting the persistence of \(H_1\) features. Identifying these loops requires careful scale selection, as noise impacts data at multiple scales, leading to false positives. Additionally, noise alters local density and distribution, creating artificial loops. Computational algorithms for \(H_1\) are sensitive to point distances, and noise perturbs these distances, causing instability. Furthermore, noise can distort the assumed low-dimensional manifold, resulting in incorrect \(H_1\) feature identification. These factors collectively make \(H_1\) more prone to noise than \(H_0\) features.

\subsection{"Ab-initio" simulation}

We have created "ab-initio" simulations of ambiguity as defined in the introduction on the sole linear representation hypothesis experiment. 
In this framework, random topics are represented as a vocabulary of normalized and orthogonal vectors in an arbitrary embedding space and datapoints are a linear combinations of those topics. For simplicity We assume uniform distribution of the topics that compose a datapoint.

\subsubsection{Simulation design}

The topics vocabulary is composed of \(N\) different topics and the arbitrary embedding space has \(D\) dimensions. We can therefore sample datapoints containing \(n\) features and build pairs of  queries and corpus datapoints. We designed 3 scenario that model the definitions of polysemy and multi-factuality and derived the persistent homologies results in various conditions for the hyperparameters (\(D,N, n_{\text{parent}},\varepsilon,\sigma_{\text{noise}} \)):

\textbf{Scenario 1 hypothesis, pure multi-factual regime}: Parent queries (topics set \(S^{\text{parent}}_{i}\)) retrieving children (topics sets \(\{S^{\text{child}}_{i,j} : S^{\text{child}}_{i,j} \subset S^{\text{parent}}_{i} \}\) ) should show strong connectivity as children inherit parent topics, as children embeddings lie in a subspace spanned by the parent's topics.

\textbf{Scenario 2 hypothesis, pure polysemic regime}: Child queries (topics sets \(\{S^{\text{child}}_{i,j} : S^{\text{child}}_{i,j} \subset S^{\text{parent}}_{i} \}\)) retrieving parents (\(\{S^{\text{parent}}_{i}\}\)) may show weaker connectivity as parent embeddings span additional topic dimensions beyond those present in the child.

\textbf{Scenario 3 hypothesis, locally multi-factual}: Mixing quasi-orthogonal children (from different parent lineages) with non-orthogonal children (from same parent) should produce intermediate scenario where topological structure varies significantly from local to global neighborhood.

Each corpus sample is composed of 50 datapoints. A detailed explanation of the algorithm can be found in appendix \ref{appendix:recursive-homology}.

\subsection{Experiment Methodology}

The challenge of evidencing consistent homology signal in an real dataset is the lack of prior information on the topics distribution in the corpus. We adopted a proxy strategy consisting of chunking queries and answers with different chunk sizes to approximate the number of topics in the query and corpus. Again, we motivate this choice with the empirical evidence that text sequences are constructed from distributions over topics where a sequence contains a set of topics and a subsequence contains a subset of those topics. It has the advantage of correlating the amount of topics in queries and answers and therefore limiting potential confounding factors in the observed topological signal. Synthetic queries generation does not guarantee such a degree of control to evidence the meaningful signals which explain why we have not chosen this popular dataset generation.

We conducted our experiments using the a set of Nobel Prize Physics lectures from 1901 to 2024, available as open-access transcripts on the Nobel Prize website. The corpus contains 14 lectures and about 250000 tokens in total. Each lecture was segmented into contiguous, non-overlapping chunks at two granularity : 250 (``fine'') and 750 (``coarse'').
This produced approximately 919 and 320 chunks respectively (Table~\ref{tab:datasets}).


\begin{table}[h!]
    \begin{center}
        \begin{tabular}{|c|c|c|}
            \hline
            \textbf{Dataset}      & \textbf{Chunk size (tokens)} & \textbf{\# chunks} \\
            \hline\hline
            \(\mathcal{D}_{750}\) & 750                          & 320                \\
            \(\mathcal{D}_{250}\) & 250                          & 919                \\
            \hline
        \end{tabular}
        \caption{Nobel-lectures corpus: chunk granularities and counts}
        \label{tab:datasets}
    \end{center}
\end{table}

We conducted semantic searches using cosine similarity metric for each model and dataset, and collected the top k = 50 retrieved answers for each query. We encoded the content of these datasets using \textit{four different embedding models}, as detailed in Table \ref{tab:models}, then stored each resulting vector embedding in separate FAISS (\cite{douze2024faiss}) vector store indexes. These models were selected due to the diversity of corpora used in their pre-training and the number of embedding dimensions. \textit{None of these models were fine-tuned on our datasets for the experiment}.

\begin{table}[h!]
    \begin{center}
        \begin{tabular}{|c|c|}
            \hline
            \textbf{Embedding model}                                                                                             & \textbf{Vector dimension} \\
            \hline\hline
            \href{https://cloud.google.com/vertex-ai/generative-ai/docs/model-reference/text-embeddings-api}{text-embedding-005} & 768                       \\
            \href{https://platform.openai.com/docs/guides/embeddings}{text-embedding-v3-small}                                   & 1536                      \\
            \href{https://platform.openai.com/docs/guides/embeddings}{text-embedding-v3-large}                                   & 3072                      \\
            \href{https://huggingface.co/tasksource/ModernBERT-base-embed}{tasksource/ModernBERT-base-embed}                     & 768                       \\
            \hline
        \end{tabular}
        \caption{Embedding models used in the experiments}
        \label{tab:models}
    \end{center}
\end{table}

The generic experimental design shown is the following:

\begin{enumerate}
    \item \textbf{Chunk formation}
          Divide the dataset \(\mathcal{D}\) into chunks of size (number of tokens)
          \(T \in \mathcal{I} = \{250,\,750\}\). 
          This produces the two test chunk sets \(\{\mathcal{C}_{T}\}_{T\in\mathcal{I}}\)
          listed in Table~\ref{tab:datasets} with chunking consistency that requests \( \bigcup_{c \in \mathcal{C}_{250}} c = \bigcup_{c \in \mathcal{C}_{750}} c = \mathcal{D}\).
    \item \textbf{Setup}
          Use one set as the corpus \(\mathcal{C}_{T} = \mathcal{C}\)
          and sample queries from the other set \(\mathcal{Q} \sim \mathcal{C}_{T'}\) as the query set \(\mathcal{Q}\).
          Generate embeddings for both \(\mathcal{Q}\) and \(\mathcal{C}\),
          index \(\mathcal{C}\) 
    \item \textbf{Retrieval}
          Perform semantic search on corpus chunks for each query and keep the top-\(k=50\) retrieved corpus chunks.
    \item \textbf{Evaluation}
          Compute persistent homology features and test their ability
          to separate ambiguous from unambiguous queries.
\end{enumerate}


\section{Results}

\subsection{Simulation}

\begin{figure*}[htbp!]
    \centering
    \captionsetup{font=footnotesize}
    \includegraphics[width=0.7\textwidth]{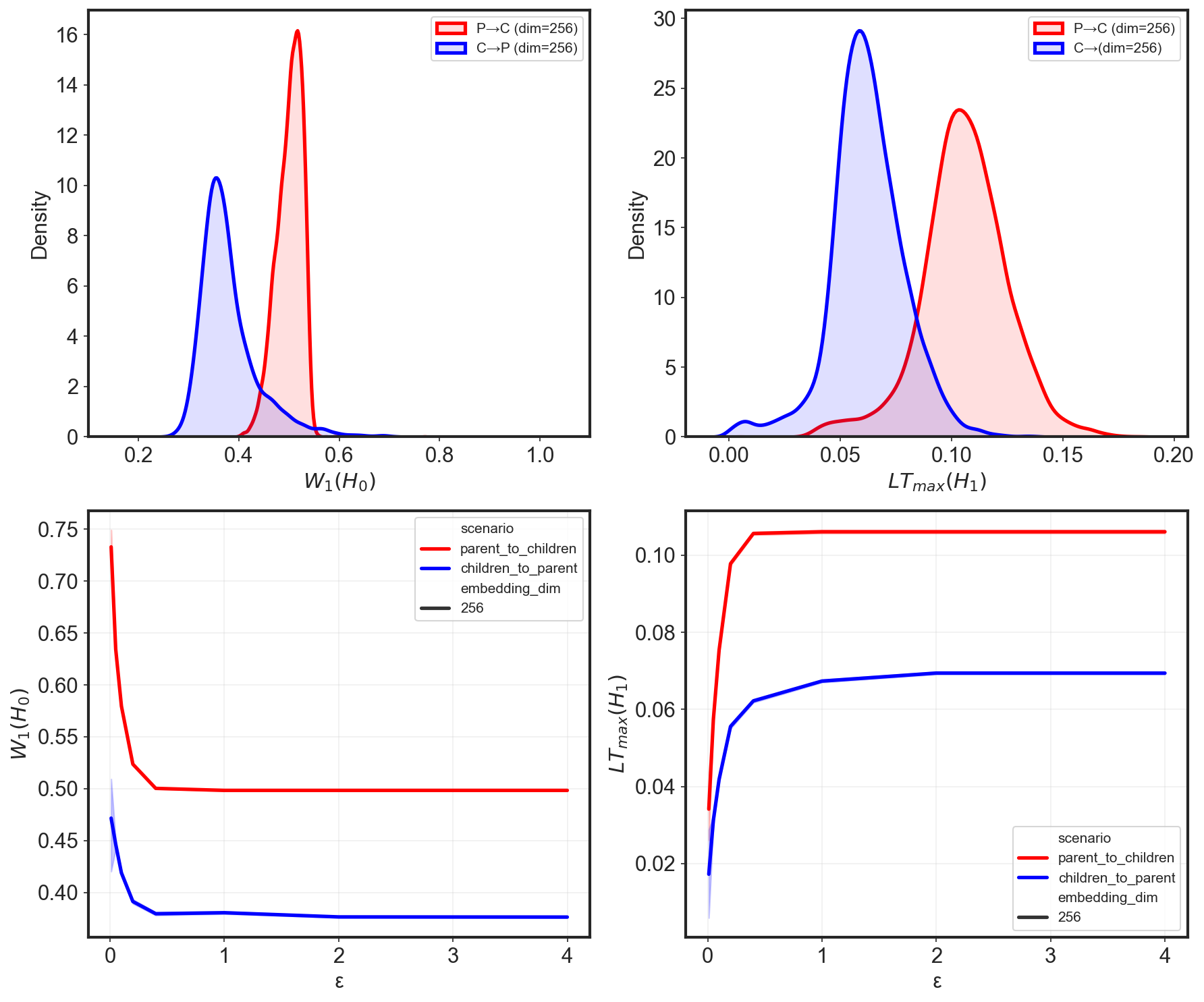}
    \caption{Top Left) Simulated \( W_{1}(H_{0})\) for scenario 1 (red) and scenario 2 (blue) with neighborhood scale \(\varepsilon=0.4\). Top Right) Simulated \(LT_{max}(H_{1})\) for scenario 1 (red) and scenario 2 (blue) with neighborhood scale \(\varepsilon=0.4\). Bottom Left) Simulated \( W_{1}(H_{0})\) for scenario 1 (red) and scenario 2 (blue) as a function of the relative neighborhood scaling factor \(\varepsilon\). Bottom Right) Simulated \(LT_{max}(H_{1})\) for scenario 1 (red) and scenario 2 (blue) as a function of the relative neighborhood scaling factor \(\varepsilon\). Simulation conditions: (\(D=256, N=64, n_{\text{parent}}=32,\varepsilon=0.4,\sigma_{\text{noise}}=0.1\))}
    \label{fig:sim_homologies}
\end{figure*}

Our "ab-initio" simulation shows a clear separation between scenarios 1 and 2 for both homology metrics \( W_{1}(H_{0})\) and \(LT_{max}(H_{1})\) (figure \ref{fig:sim_homologies} - Top) with an average separation of \(\Delta W_{1}(H_{0})\)0.2 and \(\Delta LT_{max}(H_{1}) = 0.05\). Their value as a function of the scaling of the neighborhood \(\varepsilon\) stay separated for both metrics \ref{fig:sim_homologies} - Bottom). Scenario 1 metrics stabilize around \(\varepsilon=1.0\) which corresponds to the scale where all 50 corpus datapoints are captured in the neighborhood. Scenario 2 metrics stabilize further away, around \(\varepsilon=2.0\). It is consistent with the fact that scenario 1 only consists in local neighborhood (all corpus datapoints have a subset of topics present in the query) while in scenario 2, corpus datapoints embeddings are unbounded by the query topics as they either consist in a super-set of those topics or they are uncorrelated. 

\begin{figure*}[htbp!]
    \centering
    \captionsetup{font=footnotesize}
    \includegraphics[width=0.7\textwidth]{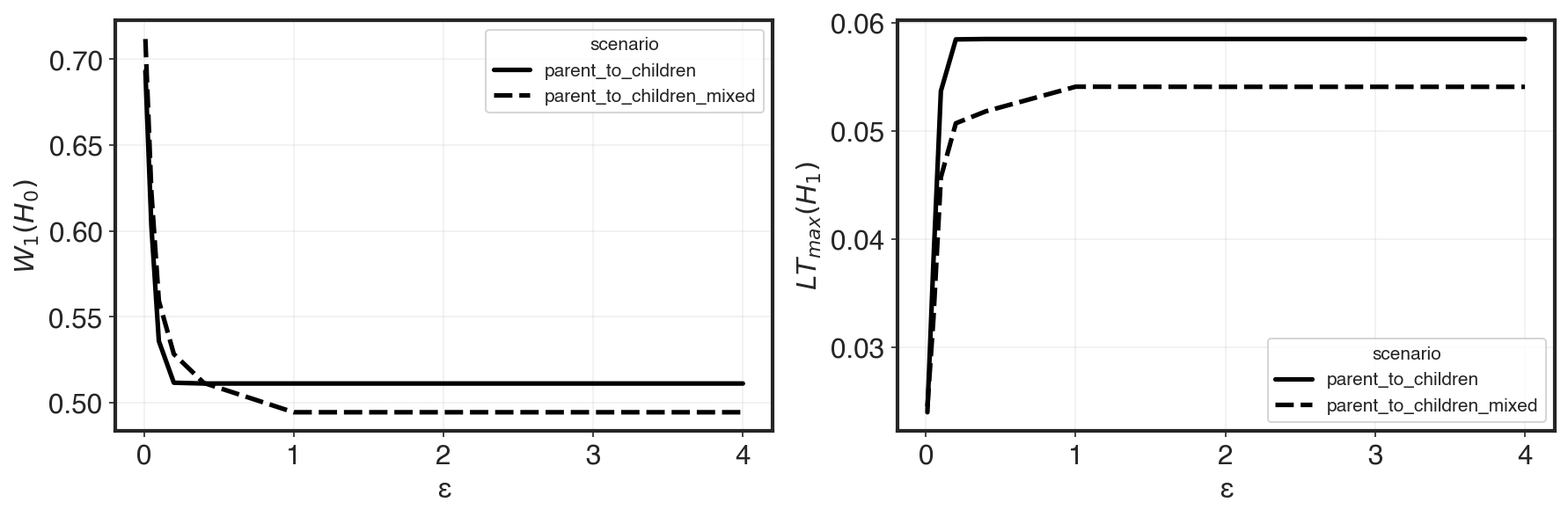}
    \caption{Left) Simulated \( W_{1}(H_{0})\) for scenario 1 (plain) and scenario 3 (dashed) as a function of the relative neighborhood scaling factor \(\varepsilon\). Right) Simulated \(LT_{max}(H_{1})\) for scenario 1 (plain) and scenario 2 (dashed) as a function of the relative neighborhood scaling factor \(\varepsilon\). Simulation conditions: (\(D=256, N=64, n_{\text{parent}}=32,\varepsilon=0.4,\sigma_{\text{noise}}=0.2\))}
    \label{fig:sim_homologies_scale}
\end{figure*}

While simulation scenarios 1 and 2 represent the two extrema of conditions, scenario 3 is a mixture of strongly and loosely correlated datapoints that should be a more faithful approximation of real world case. We can see that the \( W_{1}(H_{0})\) signal latter evolves more smoothly with \(\varepsilon\) than scenario 1 (figure \ref{fig:sim_homologies_scale}-Left) but is still separated from scenario 2 in absolute value, still allowing to determine the ambiguity regime. Similarly, absolute value for \(LT_{max}(H_{1})\) decreases and does not show as sharp increase as a function of \(\varepsilon\) (figure \ref{fig:sim_homologies_scale}-Right).
Additional results in different simulation conditions (including embedding dimensions and projections) are presented in appendix \ref{appendix:recursive-homology} and show robustness of the metrics.

Therefore, under the joint assumptions that sequences set of topics sampled from a topic distribution (\cite{grootendorst2022bertopic}) and the linear representation hypothesis to encode those topics, we demonstrate that our formal definitions of ambiguity could be translated and quantified in a metric set using persistent homologies \( W_{1}(H_{0})\) and \(LT_{max}(H_{1})\). 
In order to validate those findings in practical applications we need to evidence such signals in real world dataset. 
\newpage
\subsection{Experiment}

In our experiments, the set \(\mathcal{C}_{250}\) is initially selected as the corpus \(\mathcal{C}\), with chunks from \(\mathcal{C}_{750}\) used as the query set \(\mathcal{Q}\).
This configuration allows for the inference of query-related topics,
because at least one ``fine'' chunk in \(\mathcal{C}_{250}\) shares token sub-sequences with every query chunk drawn from \(\mathcal{C}_{750}\). this setup corresponds to the multi-factual query case (scenario 2 in our simulation).

The experimental setup is reversed by selecting \(\mathcal{C}_{750}\) as the corpus and \(\mathcal{C}_{250}\) as the queries. This corresponds to the polysemic regime (scenario 1) where the queries are generally composed of less topics than the corpus chunks.

To maintain the integrity of our experiments, we stipulate that the top retrieved chunk, denoted as \(c_{i} = [w^i_{1}, \ldots, w^i_{m}]\), must fully contain or be contained by the query chunk in both experimental configurations \((\mathcal{Q};\mathcal{C})\):
\((\mathcal{C}_{750};\mathcal{C}_{250})\) and
\((\mathcal{C}_{250};\mathcal{C}_{750})\). This ensures that the semantic coherence between the chunks is adequately maintained, allowing for accurate assessment of query ambiguity and topic distribution within the chunks.


\begin{figure*}[htbp!]
    \centering
    \captionsetup{font=footnotesize}
    \includegraphics[width=0.7\textwidth]{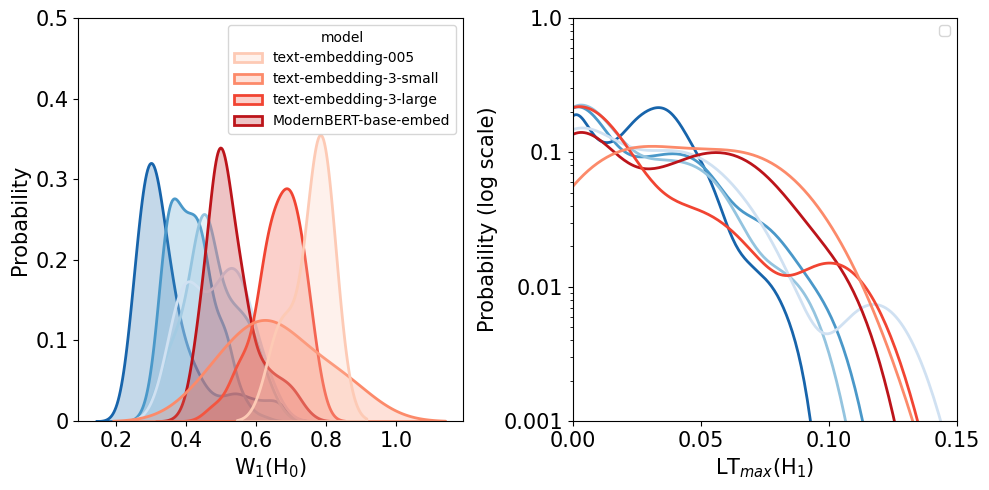}
    \caption{Left) \(W_{1}(H_{0})\) KDE plot distributions for pairs of queries-corpus \((\mathcal{Q};\mathcal{C})\) chunk sets  (\(\mathcal{C}_{250}\);\(\mathcal{C}_{750}\)) and
        (\(\mathcal{C}_{750}\);\(\mathcal{C}_{250}\)) in blue and red respectively for neighborhood scale \(\varepsilon=0.4\). Right) Max \(H_{1}\) loop lifetime KDE plot distributions for (\(\mathcal{C}_{250}\);\(\mathcal{C}_{750}\)) and
        (\(\mathcal{C}_{750}\);\(\mathcal{C}_{250}\)) in blue and red respectively with neighborhood scale \(\varepsilon=0.4\).}
    \label{fig:wasserstein_norm}
\end{figure*}

\begin{figure*}[htbp]
    \centering
    \captionsetup{font=footnotesize}
    \includegraphics[width=0.7\textwidth]{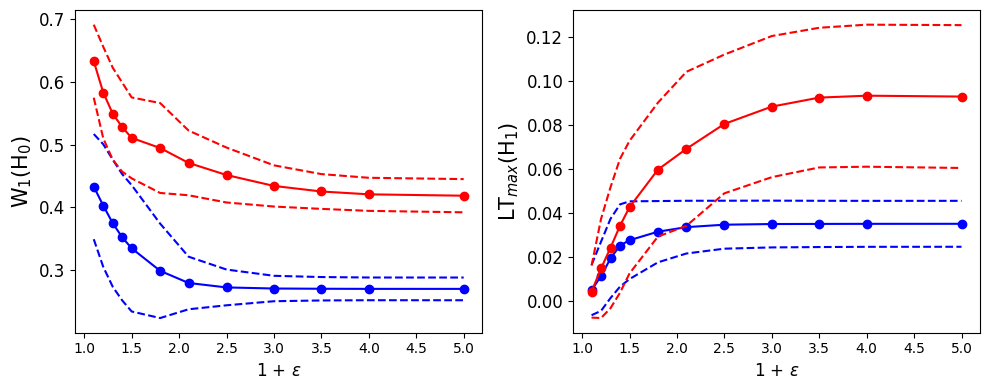}
    \caption{Left) Mean \( W_{1}(H_{0})\) (plain lines) and 25-75\% inter-quantile range (dashed lines) for  (\(\mathcal{C}_{250}\);\(\mathcal{C}_{750}\)) and
        (\(\mathcal{C}_{750}\);\(\mathcal{C}_{250}\)) in blue and red respectively as a function of the relative neighborhood scaling factor \(\varepsilon\). Right) equivalent plot for \(LT_{max}(H_{1})\). The presented data is for ModernBERT-base-embed model.}
    \label{fig:wasserstein_evolution}
\end{figure*}



In the experimental pairing of \(\mathcal{Q}=\mathcal{C}_{250}\) and \(\mathcal{C}=\mathcal{C}_{750}\), where the predefined conditions of polysemy are met, the \(H_{0}\) signal is in the region \(W_{1}(H_{0}) = 0.2-0.5\) (figure  \ref{fig:wasserstein_norm}-Left).
In contrast, the pair \(\mathcal{Q}=\mathcal{C}_{750}\) and \(\mathcal{C}=\mathcal{C}_{250}\) exhibits a shifted distribution to a higher value of \(W_{1}(H_{0})\) by a consistent \(\approx +0.2\) across all models tested. This shift is a partial indication of a more structured neighborhood. We cannot observe any obvious pattern related as a function of the embedding size of the models and all results appear in agreement with the simulation.
Similarly, as the neighborhood scaling factor \(\varepsilon\) increases, both distributions shift to lower values while maintaining the same mean distance  (figure \ref{fig:wasserstein_evolution}-Left) as in figure \ref{fig:sim_homologies}-Top. 

Same applies to \(H_{1}\) which capture the lifetime of loops existing in the data points cloud. It exhibits more persistent elements (figure \ref{fig:wasserstein_norm}-Right) as expected from our model. As the neighborhood evolves from local to global (by varying \(\varepsilon\)), the ambiguous queries also start to exhibit persistent loops (figure \ref{fig:wasserstein_evolution}-Right). Those loops can be interpreted as discontinuities that mark separation between different semantic clusters, reinforcing the picture of a pinched manifold that reflects a complex and discontinuous local embedding space surrounding the query chunks. the scaling is notably increased in the mean lifetime of unambiguous queries loops.

\section{Discussion}

\subsection{Homology encodes ambiguity}
We showed that persistent homology metrics significantly and consistently vary as a function of the query-corpus pairs conditions the query both in empirical setting on a real dataset with diverse range of embedding models as well as in an "ab-initio" simulation where data and encoding functions are abstracted and only relying on the principle of linear representation of features/topics. 

This behavior is observed across all 4 different embedding models both for \(H_{0}\) and \(H_{1}\) and in agreement with our simulation demonstrating that topological signatures of ambiguity are invariant with embedding model architecture variation and training corpus.
The agreement on \(W_{1}(H_{0})\) between simulation and ModernBERT embeddings is remarkable. It is important though to discuss the shift upward for the 3 other models. They have in common of having been trained with the Matryoshka Representation Learning loss as a regularization function (\cite{kusupati2024matryoshka}, \cite{lee2024gecko}). It optimizes the representation of a sequence of text in subspaces of decreasing dimensions all encapsulated in the same vector and that diverges strongly from the construction of datapoints in our simulation. ModernBERT, on the other hand is exempt from regularization. Despite this shift, polysemic and multi-factual regimes are well separated in all models.

In a practical case where the specifics of the embedding model are unknown, one could perform a calibration step of ambiguous vs unambiguous persistent homology values using the same methodology inspired by our real data experiment:
\begin{itemize}
    \item Identify clusters of documents.
    \item Sample documents per cluster to create multi-factual queries and embed them
    \item Run similarity search with queries \(\leftrightarrows\) corpus
\end{itemize}

Those results allows us to state that persistent homologies \(H_{0}\) and \(H_{1}\) are only functions of the polysemy of the queries and therefore encode the relative ambiguity of a query with a specific neighborhood.
Nevertheless, given the limited number of nearest neighbors in the vicinity of the query it will likely be challenging to derive statistically meaningful signals from higher order homologies.

Our findings underscore \(H_0\) and \(H_1\) homology signatures captured by \(W_{1}(H_{0})\) and \(LT_{max}(H_{1})\) as powerful tools to differentiate the two sets of query-corpus we explored. The design of our experiment aligns with our formal definition of sentence ambiguity/clarity.

\subsection{Retrieval recall optimisation}

As discussed above, absolute values of our homology metrics encode  ambiguity by defining which regime the search query is in. If the query is multi-factual it is also important to determine the optimal top k retrieval value. It is easy to identify that \(W_{1}(H_{0})\) has the same global trend in the experiment and in the simulation as the neighborhood expand but the experimental result show non monotonous behavior of \(dW_{1}(H_{0}, \varepsilon)/d\varepsilon\) resulting in the existence of different slopes. We attribute this discrepancy to the "purity" of the corpus created in the simulation versus the real world case: that is to say \textbf{all corpus datapoints} are either strongly correlated to the query (scenario 1) or loosely correlated (scenario 2) so that the distribution of degrees of correlation between the retrieved chunks and the query is bi-modal. On the contrary, in the experiment the distribution of degrees of correlation between the retrieved chunks and the query is likely more complex. Scenario 3 simulation illustrate that neighborhood heterogeneity is encoded in the homology metrics evolution as a function of \(\varepsilon\).
Therefore, what we observe here is could be used to perform hierarchical clustering of the retrieved chunks as a function of epsilon akin to HDBSCAN (\cite{McInnes2017}) and to optimize recall by dynamically updating \(k\) the number of retrieved chunks to return.

Furthermore we state that the question of recall optimization is only relevant in the case where the query is multi-factual. Indeed, a polysemic query by definition lacks of information and recalling loosely correlated datapoints would only inject noise as an output of the retrieval process and in RAG systems it can lead to generative model misrepresentation of the context and hallucination. For such a query, the best approach would be an exception handling asking for a reformulation with more context.

So far we have presented multi-factual queries simulation results for (scenario 1) and highlighted the similarities with the real world experimental results. We exhibited differences between the two settings, especially for \(LT_{max}(H_{1})\) and attributed it to "purity" of the corpus in the simulation.


\section{Conclusion and Next Steps}

In this article we have defined a formal definition of ambiguity in the context of text retrieval as the intersection of topics contained in the query and its retrieved answers candidates. We evidenced two main regimes: polysemic and multi-factual. From there, using the linear representation hypothesis to encode the topics, we constructed synthetic datasets that match the different ambiguity regimes. We evidenced that, similarly to previous works on words polysemy, the different regimes of ambiguity exhibit clear and separable persistent homology \(H_0\) and \(H_1\) fingerprints.
Finally we validated those findings with real-world dataset experiment where we isolated the same topological signatures of ambiguity in a sentence.
In addition to the discriminative utility of those signals they inform us of the geometric and topological arrangement of query sentences neighborhood which can be used as another probing tool of Large Language Models representation and structuring of information. As those signature seem independent of the language model, further investigations will explore how this relationship holds across a broader range of embedding dimensions and models, aiming to refine our understanding of semantic domain analysis in textual data. It is crucial to understand this in order to properly evaluate uncertainty of similarity metrics widely used in semantic search pipelines such as RAG (\cite{gao2024retrievalaugmented}).


\section*{Acknowledgments}
Special thanks to: Paul Gaskell and Firuza Mamedova for helpful discussions and varying levels of technical support. 

\noindent \textbf{Disclosure.} \textit{All opinions are our own and don't represent that of the employers we work and/or worked for. Results here are not meant for investment purposes, and authors cannot be held liable to any misuse of the concepts proposed here.}


\appendix\label{sec:appendix}

\newpage
\section{Geometric and Topological Foundations in Query Analysis}
\label{appendix:geom-topo}

This appendix delves into the geometric and topological underpinnings essential to our analysis, particularly focusing on the application of homology theories to discern patterns within data clusters, cycles, and cavities. Our investigation employs the Ripser library enabling us to explore complex structures within high-dimensional data spaces. The rationale behind utilizing Ripser lies in its robust computational framework for extracting topological features, pivotal for understanding ambiguities in query datasets.

\subsection{Understanding Homologies}
Homologies offer a window into the intrinsic shape of data, categorized by dimensions:
\begin{itemize}
    \item \textbf{0-Homology} focuses on clusters, identifying connected components within the data.
    \item \textbf{1-Homology} reveals loops, cycles, and tunnels, offering insights into data's cyclical structure.
    \item \textbf{2-Homology} captures spheres or cavities, indicating voids within the data framework.
    \item \textbf{3-Homology}, our study's top homology, identifies complex 3D simplexes, necessitating at least five points for construction.
\end{itemize}




\subsection{Derivation of 1-Wasserstein Norm}

The Wasserstein distance is typically used for comparing dissimilarities between distributions. More specifically it has a use case to compare topological spaces from point clouds by the means of quantifying similarities between persistence diagrams (\cite{kerber2016geometry,wasserman2016topological}) for each homology orders separately. As an example for two persistent diagrams A and B of homology \(H_0\), the general formula of the Wasserstein distance in the "giotto-tda" (\cite{tauzin2021giottotda}) package is written as:
\begin{equation}
    W_{p}(A,B) = \inf_{\gamma:A\cup\Delta\rightarrow B\cup\Delta}(\sum_{u\in A\cup\Delta}\|\ x-\gamma(u)\|_{\infty}^{p})^{1/p}
\end{equation}

where \(\Delta\) corresponds to the virtual set of diagonal points in the diagram that ensures that there is a full bijection between A and B, and \(\| . \|_{\infty}\) accounts for \(\max(|x|,|y|)\) with \((x,y)\in\mathbb{R}^{2}\) the coordinate of a point in the diagram. In our experiments we use \(W_{1}\), homology \(H_{0}\) so \(\max(|x|,|y|) \rightarrow |y|\) as all components lie on the vertical axis \((x=0,y)\) and instead of comparing two diagrams A and B we will calculate the norm for diagram A only which means that we replace B by the empty set \(\{\emptyset\}\). Finally we can discard the infimum condition as the orthogonal projection of any point on \((x=0,y)\) onto the diagonal axis \((x,y=x)\) is the function that minimizes the distances. Therefore the formula transforms into:

\begin{equation}
    W_{1}(A) = \sum_{y\in A}|\ y-\gamma_{\perp}(y)|
    \label{eq:1-wassestein}
\end{equation}

Where \(\gamma_{\perp}(y)\) is the y-component of the orthogonal projection \(\gamma(y)\) on the diagram diagonal.

\subsection{Pseudo-code for Calculating Homology features}

\begin{algorithm}[htbp]
    \caption{Calculate Homology Birth Metrics}
    \SetAlgoLined
    \KwIn{Re-normalized embeddings \(\bm{v_E}\)}
    \KwOut{Homology 0 and 1 metrics, \(W_1(H_0)\) and \(LT_{\text{max}}(H_1)\)}
    \SetKwFunction{CalculateHomologyMetrics}{CalculateHomologyMetrics}
    \SetKwProg{Fn}{Procedure}{:}{}
    \Fn{\CalculateHomologyMetrics{\(\bm{v_E}\)}}{
        Initialize Vietoris-Rips complex with the re-normalized embeddings \(\bm{v_E}\)\;
        Fit the complex to the embeddings\;
        Extract the first persistence diagram\;
        Extract homology (birth, death) ordered pairs from the diagram\;
        Compute \(W_1(H_0)\) using Wasserstein distance\;
        Compute \(LT_{\text{max}}(H_1)\) using the lifetime of the longest feature\;
    }
\end{algorithm}

Incorporating these insights enhances our understanding of the dataset's structure, offering a nuanced view of how topic diversity impacts the geometric and topological features we observe. Through this detailed analysis, we aim to provide a comprehensive framework for assessing query ambiguity, leveraging the sophisticated tools offered by persistent homologies to navigate the complexities of high-dimensional data spaces.


\section{"Ab-Initio" Simulated Persistent Homology}
\label{appendix:recursive-homology}

This appendix describes an simulation framework for analyzing persistent homology features in hierarchical cluster structures generated through orthogonal feature-based embeddings. We investigate three scenarios that model different query-corpus relationships commonly encountered in document retrieval and semantic search systems: (1) parent queries to children corpora, (2) child queries to parent corpora, and (3) parent queries to children with variable correlations. The framework systematically varies noise levels \(\sigma_{\text{noise}}\), dimension ratios (\(n_{\text{parent}}/n_{\text{child}}\)), neighborhood locality (scaling factor \(\varepsilon\)) and feature overlap to understand how topological features reflect semantic relationships.

\subsection{Data Generation: Topic Vocabulary-Based Clustering}

We generate synthetic hierarchical clusters using a vocabulary of orthogonal topic vectors. Each datapoint is constructed by sampling and summing topics from this vocabulary, then adding gaussian noise. This models document embeddings where semantic concepts (topics) are combined to create complex representations.

\subsubsection{Orthogonal Topic Vocabulary Generation}

The topics vocabulary is created using QR decomposition to ensure approximate orthogonality:

\begin{algorithm}[H]
    \caption{Generate Orthogonal Topics Vocabulary}
    \SetAlgoLined
    \KwIn{\(N\): total number of topics, \(D\): embedding dimension}
    \KwOut{\(\mathcal{F} = \{\mathbf{f}_0, \ldots, \mathbf{f}_{N-1}\}\) where \(\mathbf{f}_i \in \mathbb{R}^D\)}
    \(M \leftarrow\) random matrix \(\in \mathbb{R}^{D \times n_{\text{topics}}}\) from \(\mathcal{N}(0, 1)\)\;
    Compute QR decomposition: \(M = QR\) where \(Q \in \mathbb{R}^{D \times n_{\text{features}}}\) is orthonormal\;
    \For{\(i = 0\) \KwTo \(n_{\text{topics}} - 1\)}{
        \(\mathbf{v}_i \leftarrow Q[:, i]\) \tcp{Extract \(i\)-th orthonormal column}
        \(\text{threshold} \leftarrow \text{median}(\mathbf{v}_i)\)\;
        \(\mathbf{b}_i \leftarrow (\mathbf{v}_i \geq \text{threshold})\) \tcp{Binary mask: 1 if above median, 0 otherwise}
        \(\mathbf{f}_i \leftarrow \mathbf{b}_i / \|\mathbf{b}_i\|_2\)

        \(\mathcal{F}[i] \leftarrow \mathbf{f}_i\)\;
    }
    \KwRet \(\mathcal{F}\)
\end{algorithm}

\begin{itemize}
    \item \textbf{Orthogonality}: Topics from the vocabulary are approximately orthogonal due to QR decomposition
    \item \textbf{Arbitrary basis}: Topics are distributed across all \(D\) dimensions of the embedding space reflecting real case where there is no trivial alignment between the embedding space basis and the basis defined by the topics.
\end{itemize}

\subsubsection{Datapoint Generation with Feature Sampling}

We construct a simple 2-level hierarchy for controlled experimentation:
\begin{itemize}
    \item \textbf{Parent level}: Clusters with \(n_{\text{parent}}\) features (depth 0)
    \item \textbf{Children level}: Clusters with \(n_{\text{child}} < n_{\text{parent}}\) features (depth 1)
\end{itemize}

Each datapoint is generated by sampling a subset of features from the feature vocabulary and adding variable gaussian noise:

\begin{algorithm}[H]
    \caption{Generate Datapoints from Topics}
    \SetAlgoLined
    \KwIn{\(k\),  \(\mathcal{I}_{\text{parent}} = \{i_1, \ldots, i_{n_{\text{parent}}}\}\) or \(\{\emptyset\}\), \(\mathcal{F}\), \(n_{\text{points}}\), \(\sigma_{\text{noise}}\)}
    \KwOut{\(X \in \mathbb{R}^{n_{\text{points}} \times D}\)}
    Initialize \(X \in \mathbb{R}^{n_{\text{points}} \times D}\)\;
    \For{\(j = 1\) \KwTo \(n_{\text{points}}\)}{
        \If{ \(\mathcal{I}_{\text{parent}}=\{\emptyset\}\)}{
        Sample \(k\) topics: \(\{\mathbf{f}_{j,1}, \ldots, \mathbf{f}_{j,k}\}\) from \(\mathcal{F} \) without replacement\;
        }
        \Else{
        Sample \(k\) topics: \(\{\mathbf{f}_{j,1}, \ldots, \mathbf{f}_{j,k}\}\) from \(\{\mathcal{F}[i] : i \in \mathcal{I}_{\text{parent}}\}\) without replacement\;
        }
        Sum topics: \(\mathbf{x}_j \leftarrow \sum_{\ell=1}^k \mathbf{f}_{j,\ell}\)\;
        \If{\(\sigma_{\text{noise}} > 0\)}{
            Sample noise: \(\boldsymbol{\epsilon}_j \sim \mathcal{N}(0, \sigma_{\text{noise}})\)

            Add noise: \(\mathbf{x}_j \leftarrow \mathbf{x}_j + \boldsymbol{\epsilon}_j\)\;
        }
        Normalize: \(\mathbf{x}_j \leftarrow \mathbf{x}_j / \|\mathbf{x}_j\|_2\)\;
        \(X[j, :] \leftarrow \mathbf{x}_j\)\;
    }
    \KwRet \(X\)
\end{algorithm}

\begin{itemize}
    \item \textbf{Additive composition}: Datapoints are weighted sums of sampled topics
    \item \textbf{Controlled hierarchy}: Parent-child relationships with explicit topics count control
    \item \textbf{Topic overlap testing}: Can generate orthogonal or overlapping parent-child pairs

\end{itemize}

\subsection{Robustness of the results}

Different simulation conditions have been tested and exhibit consistent persistent homology pattern.

\begin{figure*}[htbp!]
    \centering
    \captionsetup{font=footnotesize}
    \includegraphics[width=0.7\textwidth]{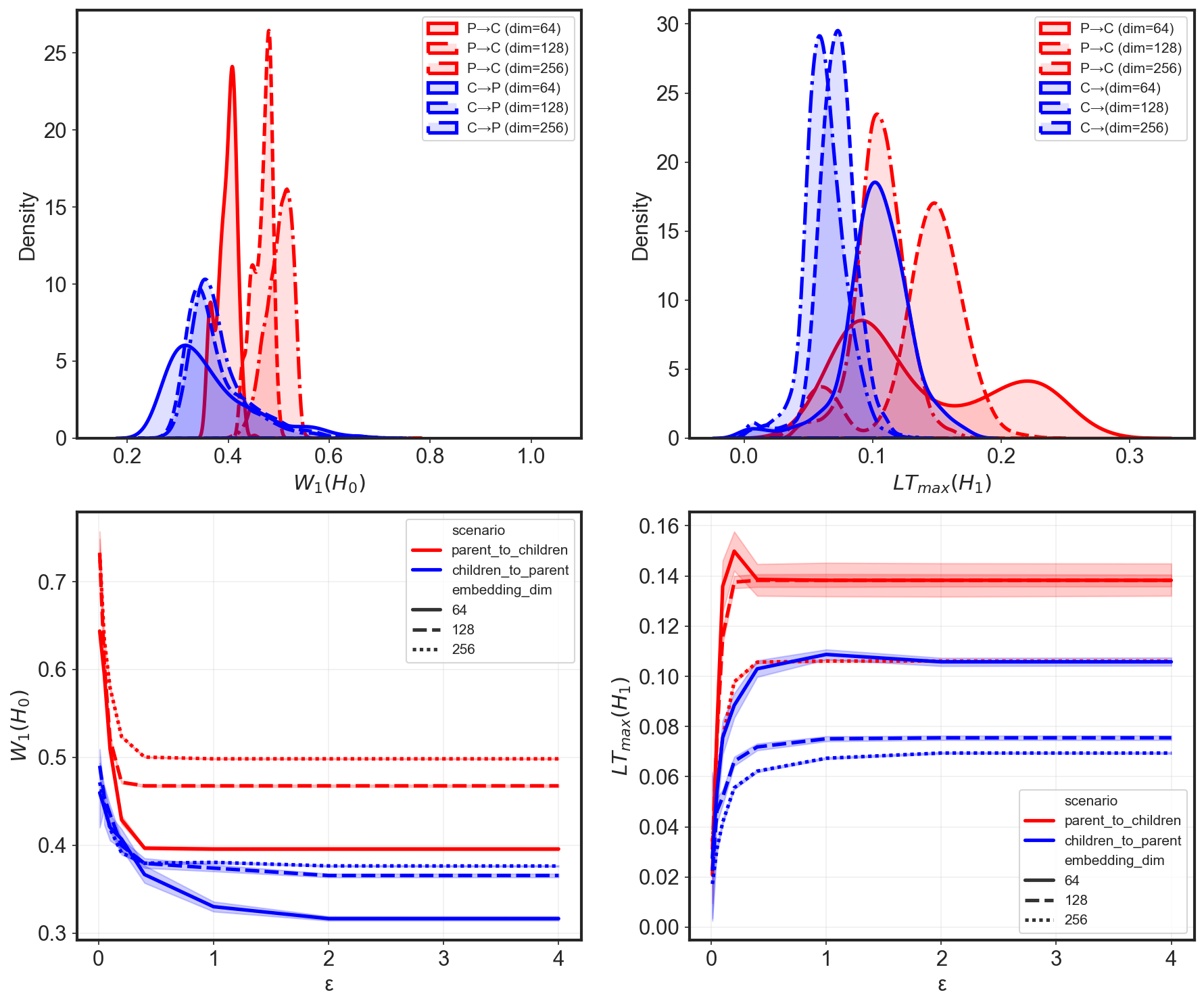}
    \caption{Top Left) Simulated \( W_{1}(H_{0})\) for scenario 1 (red) and scenario 2 (blue). Top Right) Simulated \(LT_{max}(H_{1})\) for scenario 1 (red) and scenario 2 (blue). Bottom Left) Simulated \( W_{1}(H_{0})\) for scenario 1 (red) and scenario 2 (blue) as a function of the relative neighborhood scaling factor \(\varepsilon\). Bottom Right) Simulated \(LT_{max}(H_{1})\) for scenario 1 (red) and scenario 2 (blue) as a function of the relative neighborhood scaling factor \(\varepsilon\). Varied simulation conditions: \(\{(D=64, N=16, n_{\text{parent}}=12),(D=128, N=32, n_{\text{parent}}=16), (D=256, N=64, n_{\text{parent}}=32)\}\), Fixed simulation conditions: \(\varepsilon=0.4,\sigma_{\text{noise}}=0.1\)}.
    \label{fig:sim_homologies_all_d}
\end{figure*}

\begin{figure*}[htbp!]
    \centering
    \captionsetup{font=footnotesize}
    \includegraphics[width=0.7\textwidth]{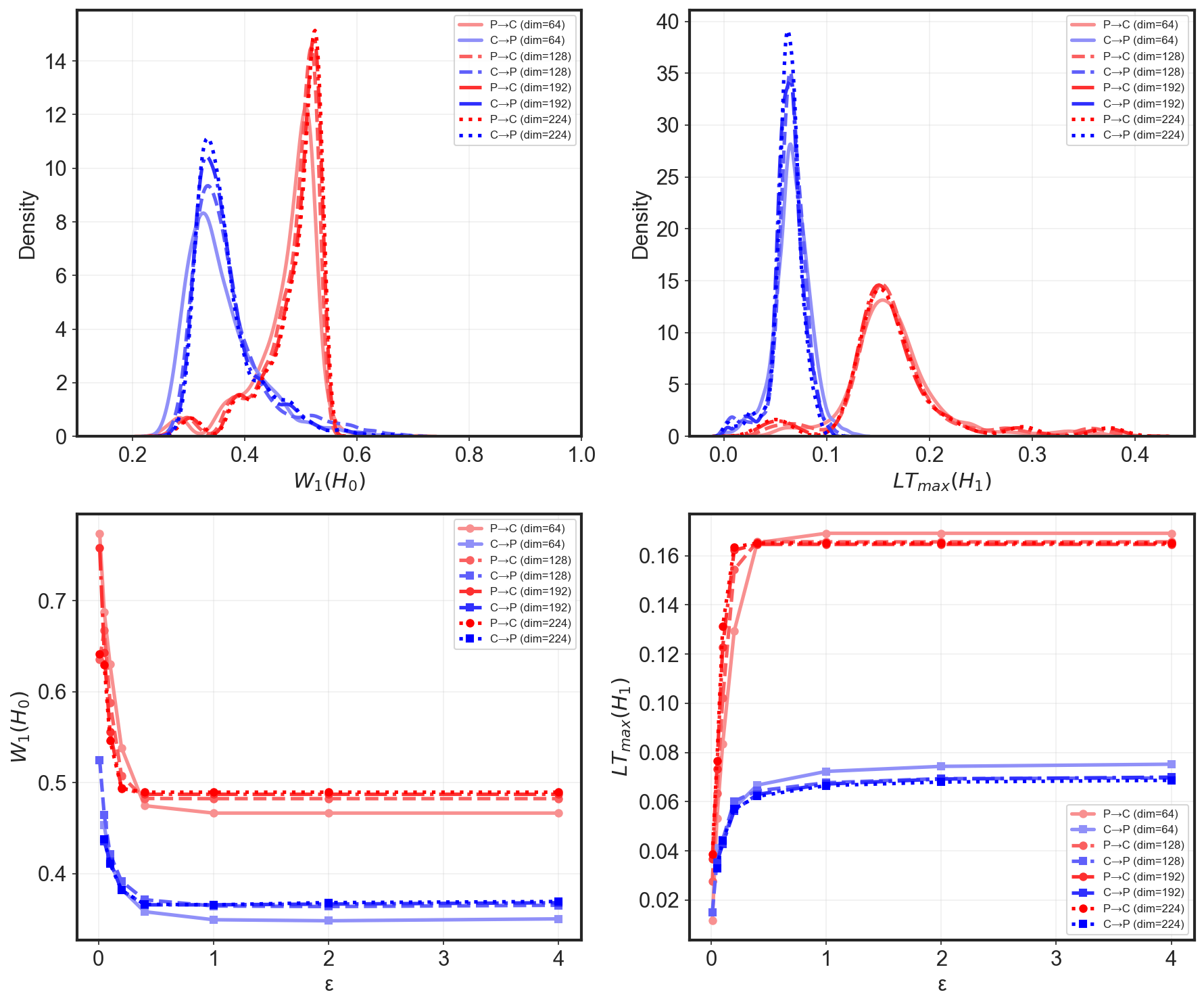}
    \caption{Projecting query and corpus embeddings from 256 dimensions to \(\{224,192,128,64\}\) dimensions with random projection matrix. Starting Simulation conditions: (\(D=256, N=64, n_{\text{parent}}=32,\varepsilon=0.4,\sigma_{\text{noise}}=0.01\))}
    \label{fig:homologies_with_projection}
\end{figure*}

We also performed an additional test of robustness by applying dimensionality reduction to our embedding vectors. According to the Johnson-Lindenstrauss lemma, a random projection to a lower dimension space \(D'<D\) preserves the distances between the datapoints and we expect the homology signal to be stable under that transformation. We report here that our simulation homology signal is invariant with respect to such a random projection (figure \ref{fig:homologies_with_projection})which reinforce our claim that it depends on the relative topics encoding of datapoints and not the absolute dimensionality of the embedding space.

\newpage
\newpage
\bibliographystyle{plainnat}
\bibliography{blowfish_v3} 

\end{document}